\documentclass[pdflatex,sn-mathphys-num]{sn-jnl}

\usepackage{graphicx}%
\usepackage{multirow}%
\usepackage{amsmath,amssymb,amsfonts}%
\usepackage{amsthm}%
\usepackage{mathrsfs}%
\usepackage[title]{appendix}%
\usepackage[T1]{fontenc}
\usepackage{xcolor}%
\usepackage{textcomp}%
\usepackage{manyfoot}%
\usepackage{booktabs}%
\usepackage{algorithm}%
\usepackage{algorithmicx}%
\usepackage{algpseudocode}%
\usepackage{listings}%

\theoremstyle{thmstyleone}%
\theoremstyle{thmstyletwo}%

\theoremstyle{thmstylethree}%

\begin{document}

\title[Article Title]{Quantifying Full-Body Immersion}


\author[]{\fnm{Alihan} \sur{Bakir}}\email{alihan.bakir@epfl.ch}

\author[]{\fnm{Ekrem} \sur{Y\"{u}ksel}}\email{ekrem.yuksel@epfl.ch}
\equalcont{These authors contributed equally to this work.}

\author[]{\fnm{Fabio} \sur{Zuliani}}\email{fabio.zuliani@miros.work}
\equalcont{These authors contributed equally to this work.}

\author[]{\fnm{Neil} \sur{Chennoufi}}\email{neil.chennoufi@miros.work}

\author[]{\fnm{Francesco} \sur{Bruno}}\email{francesco7.bruno@mail.polimi.it} 

\author*[]{\fnm{Jamie} \sur{Paik}}\email{jamie.paik@epfl.ch}

\affil{\orgdiv{Reconfigurable Robotics Lab}, \orgname{EPFL}, \orgaddress{\street{MED 1 2313 Building MED Station 9}, \city{Lausanne}, \postcode{1015}, \state{Vaud}, \country{Switzerland}}}


\abstract{Humanity is at the forefront of yet another digital revolution, where the lines between real and virtual worlds are dissolving, reshaping how we perceive and interact with our surroundings. In this context, we introduce a transformative paradigm for immersive virtual experiences centered around whole-body kinetic interactions. Our approach redefines immersion through three distinct levels: audio-visual immersion, capturing sensory realism; physical immersion, delivering haptic feedback; and full-body immersion (FBI), where dynamic bodily interaction integrates seamlessly with virtual environments. At the core of this innovation lies a scalable, distributable platform based on modular robotic surface units inspired by the adaptive designs of nature. These units enable the rendering of immersive environments at any scale, from intimate personal experiences to expansive multi-user settings, dynamically adapting to interactions in real-time. The modular system distributes force, shape, and motion feedback throughout entire spaces, replicating the physical characteristics of the environment and enabling new depth of engagement through FBI. By combining scalability, adaptability, and dynamic physical engagement, this framework bridges the gap between real and virtual worlds. It offers an unprecedented level of immersion where users can engage their entire bodies in symbiotic interactions with the virtual space. This work not only advances immersive technology but also redefines how humans and virtual environments coexist, setting a foundation for a new era of human-environment synthesis.}




\maketitle

\makeatletter
\newcommand{\printfontsize}{Current font size: \f@size\ pt}
\makeatother

\section{Introduction}\label{section:Intro}

Immersive devices are fundamentally transforming the way humans interact digitally, offering the promise of experiences that merge the physical and virtual worlds seamlessly. Achieving this, however, requires overcoming significant challenges: how to create environments that feel as intuitive and natural as the real world while enabling scalable and collaborative interactions. Addressing these challenges, our framework advances immersion through a layered approach that integrates audio-visual immersion, haptic immersion, and a novel concept: full-body immersion (FBI). Together, these layers establish a framework that engages multiple sensory channels while extending physical interaction into the virtual environment in a structured and quantifiable way.

Dynamic, adaptive environments capable of interacting with users across multiple levels form the core of this framework. \textit{Audio-visual immersion} lays the groundwork by anchoring users cognitively in virtual spaces, employing advanced techniques like spatial audio \cite{fabio01,fabio02} and real-time rendering of high-resolution visuals \cite{fabio03,fabio16, alihan07,alihan04,alihan13}. Building on this, \textit{haptic immersion} introduces localized tactile feedback, utilizing tools like gloves and wearables to simulate texture and resistance \cite{fabio04,fabio05,fabio17,alihan14, alihan24, alihan27}. While effective, these layers are inherently limited in their ability to fully replicate the experience of interacting with a physical environment, particularly at scale \cite{alihan08,alihan09,alihan10,alihan11,alihan12}.

Achieving immersion in interactive environments presents several fundamental challenges. First, the lack of standardized quantitative metrics makes it difficult to evaluate and compare immersive systems in a reproducible manner. Second, scalability requires architectures that support distributed physical interaction without compromising coherence across the environment. Third, coordinated control of multiple physical modules demands integration of local responsiveness with global synchronization. Overcoming these challenges requires a combination of robust quantification methods, modular hardware architectures capable of distributed physical interaction, and hierarchical control strategies that integrate local real-time responsiveness with centralized high-level decision-making.

To address these limitations, we propose MIROS (Multi-scale Interactive Reconfigurable Origami Surface) platforms, an adaptable and scalable modular system designed to enable FBI. Each MIROS platform, featuring three degrees of freedom (DoF), functions as a parallel platform with actuated legs that distribute load while maintaining structural resilience \cite{alihan20, alihan21, alihan23, alihan25, alihan26}. Drawing inspiration from natural systems \cite{fabio06,fabio07,fabio18}, these modules provide a versatile and robust foundation for creating interactive, immersive spaces of varying sizes. Modular systems like this have demonstrated transformative potential in applications ranging from robotics to adaptive architectural design \cite{fabio08,fabio09, alihan29}. For example, a set of platforms can recreate a remote interaction environment within a room while enabling the teleoperation of a distant robot, with physical feedback rendered locally through the platforms.

\begin{figure}
    \centering
    \includegraphics[width=1\linewidth]{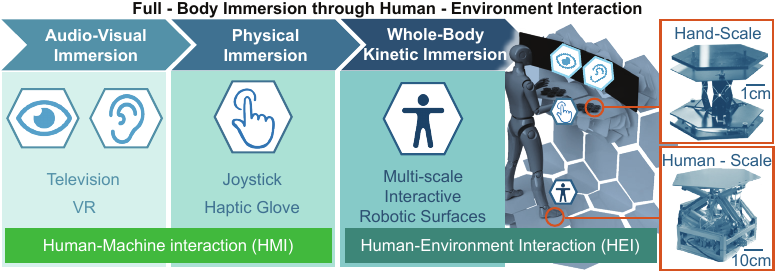}
    \caption{\textbf{Multi-level immersion framework.} This work expands on the existing work on immersion in virtual environments which started with audio and visual information with television and more recently VR headsets. Physical immersion was brought first with a joystick which allowed to give physical information and get feedback through hand gestures which was later expanded to haptic gloves concepts in the framework of human-machine interaction. Our vision brings the immersion to a new level through FBI, enabled by a distributed and scalable framework of interactive robotic surface modules that support kinetic whole-body experiences. The figure on the right shows an immersive station made with multi-scale inteactive surface modules to bridge the whole-body sensations to a virtual world.}
    \label{fig:01}
\end{figure}

Interaction is governed by a hierarchical control framework that integrates real-time local responsiveness with centralized coordination, enabling synchronized feedback and consistent data propagation across the system \cite{fabio10,fabio11}. This architecture supports scalability and reliability in both single- and multi-user scenarios \cite{alihan01,alihan02,alihan15,alihan16}. As a result, modular surfaces can dynamically adapt to user movements and distribute force and shape feedback across the space, enabling simultaneous interaction by multiple users. This supports collaborative applications such as remote design, shared virtual prototyping, and adaptive educational simulations \cite{fabio13,alihan03,fabio23}.

This work builds on interdisciplinary advances in robotics, human–machine interfaces, and adaptive systems. Innovations in modular robotics, sensorized materials, and real-time control provide the foundation for environments that replicate physical properties while dynamically adapting to user behavior \cite{alihan05,alihan06,alihan17}. We introduce a method to quantify Full-Body Immersion (FBI) based on measurable physical engagement between users and the environment, including body–platform dynamics. Validated through experimental and real-world data, the system addresses challenges in scalability, responsiveness, and multi-user integration, demonstrating robust performance and extending immersive interaction across physical and virtual domains.

\section{Results}\subsection*{Kinetic Interaction System Architecture}\label{section:System_architecture}

Physical realism plays a crucial role in immersive experiences, requiring the reproduction of not only visual input but also the dynamic forces acting on the body in virtual environments. We achieve this by mapping real-world motion into virtual representations. However, current haptic devices, despite offering high DoF in feedback \cite{fabio04}, intuitive interaction \cite{fabio17}, and stable performance \cite{alihan08}, generally lack standardized, quantifiable, and parameterized evaluation methods. Most studies rely heavily on user feedback, typically in the form of surveys, to assess performance. How these systems are quantified, combined, and compared remains inconsistent and often case-specific \cite{alihan09,alihan10,alihan11,alihan12}. Notably, none of the existing literature provides a quantifiable measure of FBI.

\begin{figure} [h]
    \centering
    \includegraphics[width=1\linewidth]{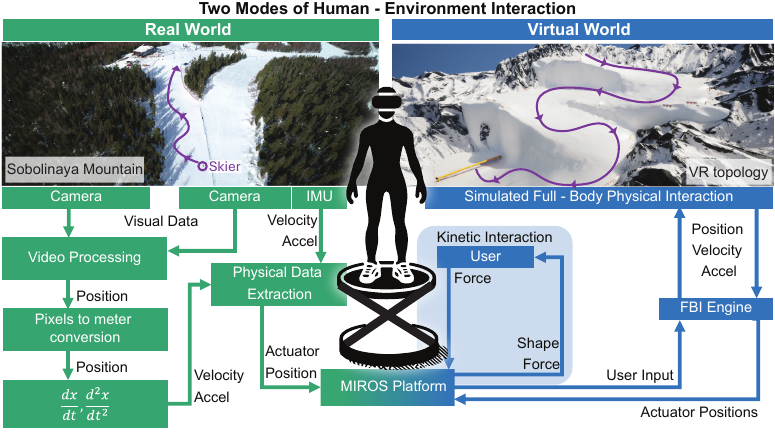}
    \caption{\textbf{Overview of the system design architecture with low and high level control loop for acceleration tracking.} The motion in real-world skiing (left) is tracked by collecting the data through video analysis to extract positional data, which is converted to acceleration. Similarly, acceleration data can be extracted from simulation environment (right), where the virtual avatar's motion is synchronized with the platform. The control system (bottom-right) uses nested PID controllers to process acceleration and leg angle errors, adjusting the platform's orientation through actuator inputs. The measured acceleration on the platform and data from two separate source (simulation and video) are compared.}
    \label{fig:02}
\end{figure}

To address the lack of a standardized metric, we introduce a scalable architecture composed of the control framework and the MIROS platform. At the full-body scale, the platform uses a metal load-bearing structure to support the dynamic load of a standing user, with an actuation bandwidth of $20 \: Hz$, peak platform velocity of $120 \: mms^{-1}$, and actuator torque capacity of $892 \: Nm$. The same transmission kinematics govern platforms from millimetre-scale prototypes to the $\sim 30 \: cm$ human-scale platform, while actuation, sensing, and structural materials scale with the load regime (details in Section \ref{sec:scalability}). Millimetre-scale modules use FR-4 rigid elements, miniature displacement sensors and Kapton folding joints, with compatibility for piezoelectric actuation and force sensing. Needle- to hand-scale modules use FR-4/Kapton laminated structures, where Kapton forms the flexural joints, and hand-scale modules use BLDC direct-drive actuation with embedded torque, displacement, and position sensing. Full-body modules use aluminium 5083 and 7075 structural components, metal ball-bearing joints, high-torque electromechanical actuation, encoder feedback, torque sensing, and user-force sensing. Joint-level motor controllers close the actuator loops using encoder feedback and torque/current measurements, while the high-level controller provides position, velocity, or force references, and synchronizes sensing, feedback rendering, and user interaction. Because the kinematics are scale-independent, the same framework drives a single full-body platform, one platform per leg, or decoupled fingertip- to full-body-scale interfaces.

This architecture supports two operating modes, feedback rendering and interactive mode, as shown in Fig. \ref{fig:02}. In feedback rendering, cameras or IMUs record real motion, and the controller converts it into position, velocity, and acceleration references rendered through the platform; for example, skiing acceleration is reproduced by preserving vertical acceleration and mapping longitudinal and lateral acceleration to pitch and roll. In interactive mode, the user applies force to the platform to control a virtual character on skis or a boat, while the controller computes the resulting environment dynamics and renders them back through MIROS. Thus, the two modes separate the robot’s ability to reproduce physical events from its ability to measure active user intention for environment interaction.

We use this architecture to eliminate the reliance on subjective and task-specific evaluation methods, introducing a quantitative, platform- and task-agnostic immersion index that captures the level of FBI. This index enables direct comparison of immersion across different tasks and interaction architectures.

\subsection*{Quantifying FBI: Real vs Virtual Performance}\label{section:Quantifying_FBI}

To achieve FBI, we implemented the MIROS platform with three DoF into the proposed system, designed to replicate the motion and acceleration experienced in real-life experiences. Although the limited DoF restricts a full one-to-one mapping of real-world motion, the platform delivers accurate feedback based on real data. This enables examination of how platform-induced motion influences user behavior. During the skiing simulation, the framework preserves vertical acceleration directly, while longitudinal and lateral accelerations map to pitch and roll through inertial equivalence. Visual feedback maintains fidelity within the platform’s actuation limits despite the absence of yaw and other residual motion components. In the boat simulation, direct roll, pitch, and heave motion of the boat drives the platform, as acceleration from sudden motion is negligible due to the slow and gradual dynamics.

\begin{equation}
\mbox{immersion\ index\ (\%)} = \frac{\mbox{intersecion\ area\ of\ gyration\ circles}}{\mbox{union\ area\ of\ gyration\ circles}} * 100\ (\%)
\label{eq:1}
\end{equation}

To quantify motion differences across setups, we extracted the interquartile ranges (IQRs) of activity-specific submetrics. These submetrics minimize sensitivity to noise and irrelevant motion because they are defined by the task and the dominant body segments involved, as detailed in Section \ref{Section:Methods}. The IQR values (Q3–Q1) construct a radar chart to form a polygon representing the motion profile for each condition: real-world ground truth, simulation without physical feedback, and the FBI environment with the MIROS platform. The centroid and radius of gyration of each polygon define equivalent gyration circles. Treating these circles as Venn diagrams, we calculated the intersection-over-union area relative to the ground truth. The immersion index is the percentage of intersection-over-union area, as Equation \ref{eq:1} defines, ranging from 0\% (no overlap) to 100\% (complete alignment).

\begin{figure}[H]
    \centering
    \includegraphics[width=1\linewidth]{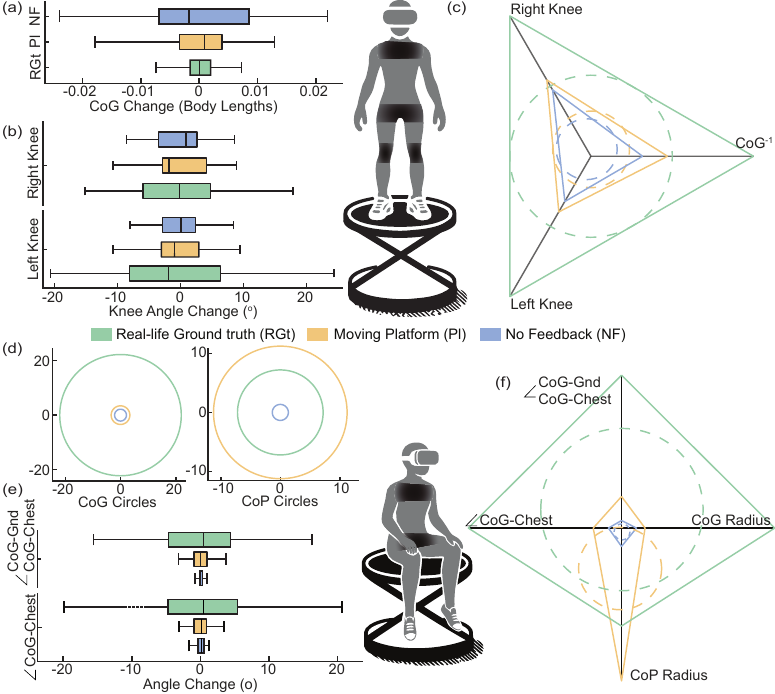}
    \caption{\textbf{Data shows the experiment results and required charts to calculate immersion index for skiing and boating.} The top half of the figure corresponds to the skiing experiments, while the bottom half corresponds to the boating experiments. Each set includes data from the corresponding real-world activity as ground truth (RGt), the simulation environment with the MIROS platform providing physical feedback (Pl), and the simulation environment without physical feedback, where the user provides input only (NF). (a) Change of CoG range of motion recorded for real-skier and user in two seperate conditions: on stationary force plate with no feedback and on the platform. (b) Range of motion in right and left knee recorded for the same three conditions. (c) The radar chart for generating the corresponding polygons and their gyration circles required for calculating the immersion ratio. (d) Range of CoG sway (standing) and center-of-pressure (CoP) variation (seated). (e) Angle between CoG–chest and CoG–feet vectors (standing) and range of CoG–chest vector angle (seated). (f) Radar chart used to compute the immersion index via gyration circles of the resulting polygons.}
    \label{fig:03}
\end{figure}

Two set of experiments quantify the level of immersion and verify the immersion index: skiing involves high acceleration dynamics and boating involves lower acceleration changes. In each experimental set, we first evaluate the performance of the platform to follow the required signal; the detailed validation of this tracking performance is in the Section \ref{Section:Methods}. Following this performance verification, we conducted a skiing experiment using a ski simulator, where the interaction occurs while user is standing and assumes symmetric movement of both legs—allowing the use of a single MIROS platform. The boat experiment includes a boat simulation, and during this experiment we evaluated both sitting and standing cases to cover the different postural configurations typically involved in driving a boat.

The simulation engine computes the avatar’s motion and corresponding acceleration to generate real-life force cues in the virtual environment. The platform reproduces these cues by dynamically adjusting roll and pitch angles, thereby reorienting the gravitational vector in the platform reference frame. Using inverse kinematics and a closed-loop acceleration controller, the framework computes the required platform orientation from the avatar’s acceleration vector and transmits the corresponding signals to the actuators. Figure \ref{fig:M3} in Section \ref{Section:Methods} details the acceleration control performance of the platform.

During skiing, the platform induces body movements that more closely resembles real-life skiing compared to a stationary force plate setup, where the simulation received user input without providing physical feedback. We analyze the body’s center of gravity (CoG) displacement and knee flexion angles. Figure \ref{fig:03} shows the user on the MIROS platform exhibits larger knee angle variations and more stable CoG trajectories, consistent with skilled skiing behavior characterized by large leg motion to keep a low, stable CoG \cite{alihan18,alihan19}.

For boating, the motion of the user resembles the real-life ground truth more closely while the user receives feedback from the platform compared to the user on the ground and not receiving any physical feedback. During this experiment we analyzed multiple submetrics: CoG sway motion and the angle between CoG-chest \& CoG-feet vector while standing as well as change of the center of pressure and CoG-chest vector angle while sitting. Submetric selection and decision process detailed in the Section \ref{Section:Methods}. Users on the platform exhibit larger CoG sway and greater relative motion between chest and legs while standing, and larger center-of-pressure shifts and greater deviation in the CoG–chest angle while sitting, as in Figure \ref{fig:03}. The user motion on the platform aligns with the ground truth (real-life boat driving data).

For skiing, the stationary force plate condition yields an immersion index of 14\%, whereas the MIROS platform increases to 23\%, indicating greater postural alignment with real skiing dynamics under physical feedback. As the boating experiments done with a controller, it shows 1\% immersion index for stationary case and 20\% when the moving platform is introduced. This shows that the effect of the moving platform is consistent across both activities.

\section{Discussion}\label{sec12}

We introduce a quantitative definition of FBI, the immersion index, based on measurable comparison between real-world ground truth and whole-body motion in a virtual environment. The ground truth is extracted from real-world video and terrain maps, capturing position, velocity, acceleration, and topology. The proposed immersion index evaluates the geometric similarity of motion variability using gyration-circle overlap derived from submetrics relevant to the task. This formulation provides a task- and platform-agnostic immersion index based on human motion.

Across both dynamic experiences (skiing and boating), full-body haptic feedback from the MIROS platform increases the immersion index relative to the no-feedback condition, indicating closer alignment with real-world human motion. Despite limited degrees of freedom and imperfect acceleration magnitude matching, the platform preserves the directional consistency of acceleration vector between real and virtual conditions. These results support the proposed immersion index as a quantitative measure of similarity between whole-body motion in real-world and virtual world. The increase in immersion index under platform feedback demonstrates that the index captures the changes in user motion induced by the system. While validated on a 3-DoF platform, the formulation is independent of the underlying hardware and can be applied to systems with higher degrees of freedom.

The proposed index captures statistical variability of task-specific motion over time rather than enforcing trajectory-level correspondence between ground truth and virtual experience. Submetric selection depends on the task and follows the criteria defined in Section \ref{Section:Methods}. The experiments were conducted using a single-module configuration across two task domains, demonstrating the effectiveness of the proposed index under minimal hardware conditions. Extension to multi-module and distributed multi-scale platforms provides a pathway to broader interaction scenarios and increased immersion with the proposed framework.

\clearpage
\section{Methods}\label{Section:Methods}

To develop a scalable, distributable and fully immersive interactive system, we designed and validated a modular robotic surface capable of real-time adaptation to human movement and force input. This section describes the computation of gyration circles and the immersion index, the scalable design and multi-material manufacturing methodology, platform performance evaluation, and the selection of submetrics. The approach ensures that the system can dynamically accommodate a wide range of applications, from localized haptic feedback to large-scale immersive environments.

\subsubsection*{Gyration circles and immersion index}

\begin{figure}[b]
    \centering
    \includegraphics[width=1\linewidth]{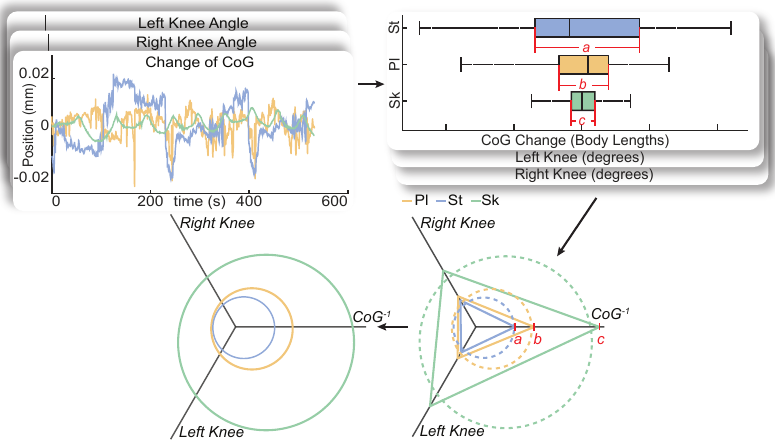}
    \caption{\textbf{The Flow of Extracting the Submetric Data for Immersion Index.} Time-series motion signals are reduced to activity-specific submetrics, quantified by interquartile ranges (IQR), and mapped onto radar polygons. Each polygon is converted to a gyration circle, and immersion is defined as the intersection-over-union area relative to the ground-truth condition. The time-series data shown here is collected during the ski simulation environment; however, the same procedure applies to boating and other activities.}
    \label{fig:M1}
\end{figure}

\begin{algorithm}
\caption{Immersion Ratio Computation using Motion Variables}
\begin{algorithmic}[1]
\Require Q1 and Q3 values for $n$ motion variables per condition
\Ensure Immersion ratio for comparison with real skier benchmark

\ForAll{conditions $\in$ \{Real Skier, Moving Platform, Stationary Force Plate\}}
    \ForAll{variables $i \in \{1, \dots, n\}$}
        \State Compute interquartile range: $IQR_i \gets Q3_i - Q1_i$
        \State Assign angle $\theta_i \gets \frac{2\pi(i-1)}{n}$ on radar chart
        \State Convert to Cartesian coordinates: \\
            \hspace{1em} $x_i \gets IQR_i \cdot \cos(\theta_i)$ \\
            \hspace{1em} $y_i \gets IQR_i \cdot \sin(\theta_i)$
    \EndFor
    \State Compute polygon centroid: \\
        \hspace{1em} $x_c \gets \frac{1}{n} \sum_{i=1}^{n} x_i$, \quad $y_c \gets \frac{1}{n} \sum_{i=1}^{n} y_i$
    \State Compute moment of inertia: \\
        \hspace{1em} $I \gets \sum_{i=1}^{n} ((x_i - x_c)^2 + (y_i - y_c)^2)$
    \State Compute polygon area $A$ using shoelace formula
    \State Compute radius of gyration: $r \gets \sqrt{I / A}$
\EndFor

\ForAll{comparison pairs $\in$ \{(Moving Platform, Real Skier), (Stationary Force Plate, Real Skier)\}}
    \State Let $(x_c^1, y_c^1, r_1)$ and $(x_c^2, y_c^2, r_2)$ be centroids and radii
    \State Compute center distance: $d \gets \sqrt{(x_c^1 - x_c^2)^2 + (y_c^1 - y_c^2)^2}$
    \State Compute intersection area $A_{\text{inter}}$ using circle–circle intersection formula
    \State Compute union area: $A_{\text{union}} \gets \pi r_1^2 + \pi r_2^2 - A_{\text{inter}}$
    \State Compute immersion ratio: \\
        \hspace{1em} $\text{Immersion} \gets \frac{A_{\text{inter}}}{A_{\text{union}}} * 100$
\EndFor

\end{algorithmic}
\end{algorithm}

To quantitatively assess motion similarity across all experimental conditions, a generalizable algorithm computes an immersion index based on geometric comparison of motion variability. The process begins by extracting interquartile range (IQR) values from box plots for a set of motion variables. For one-dimensional submetrics, variability is represented by the IQR. For two-dimensional submetrics, such as CoG sway motion and center-of-pressure variation, the data is represented in the dominant plane for that submetric (CoG sway motion is represented in $xy$ plane) and a minimum enclosing circle is computed; the radius of this circle defines the corresponding value used in the radar chart. Using the minimum enclosing circle for 2D submetrics follows the same principle as representing trajectory variability through a confidence ellipse fitted to motion data (e.g., center-of-pressure sway ellipse) as in \cite{alihan28}. The radius of the enclosing circle and IQR values constitute a radar chart forming a polygon that captures variability in each condition. Conversion to Cartesian coordinates enables computation of the centroid and radius of gyration, defined as the square root of the moment of inertia about the centroid divided by the polygon’s area. Each motion profile translates to a gyration circle in 2D space. As in Figure \ref{fig:M1}, the intersection-over-union between these circles is computed, treating them as Venn diagrams, with the real-life skier’s circle serving as the benchmark. This geometric abstraction ensures that the immersion index reflects overall motion structure rather than individual variable magnitudes. The resulting immersion ratio lies in $[0\%, 100\%]$, where 0\% indicates no overlap and 100\% denotes identical motion variability.

\subsubsection*{Submetrics}

Submetrics are task-specific motion descriptors used to compute the immersion index. Each submetric represents a measurable mechanical variable that captures a dominant component of body motion during a given activity. Examples include CoG displacement, knee flexion range, chest inclination, center of pressure while sitting, or the angle between CoG–chest and CoG–feet vectors. Submetrics are selected according to three principles:

\begin{itemize}
\renewcommand{\labelitemi}{$\circ$}
    \item \textit{Relevance:} Each submetric must represent a mechanically dominant feature in the human body during the activity. For example, knee flexion is critical in skiing, whereas trunk inclination and pressure distribution while sitting are more prominent in boating.

    \item \textit{Independence:} Submetrics should capture distinct aspects of motion and avoid redundancy. Correlated variables describing the same mechanical behavior are not included simultaneously.

    \item \textit{Robustness:} Submetrics must be derived from stable motion features and be minimally influenced by incidental body movement that is not related to the task.
\end{itemize}


At least three submetrics are necessary to construct a polygonal representation in radar space, enabling geometric comparison across conditions. For each submetric, variability is quantified using its interquartile range ($IQR = Q3 - Q1$), ensuring robustness to outliers. These IQR values form the axes of a radar chart, producing a polygon that represents the motion profile for a given condition (real-world ground truth, simulation without feedback, or FBI). For CoG sway and center-of-pressure (CoP) change, which are defined in the 2D $xy$ plane, variability is computed using the radius of the minimum enclosing circle, and this radius is used in place of the IQR in the immersion calculation.

\subsubsection*{Platform performance}

\begin{figure}
    \centering
    \includegraphics[width=1\linewidth]{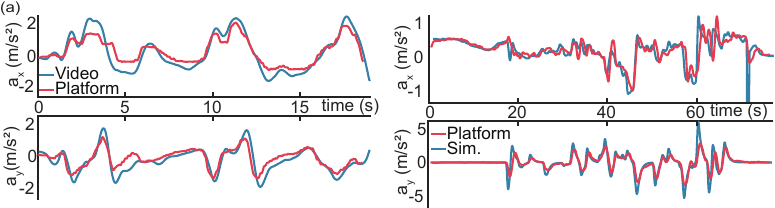}
    \caption{\textbf{The System's Ability to Replicate Real-World Forces.} Performance of the MIROS platform in replicating acceleration vectors calculated from a video, demonstrating its capability to recreate real-world forces for immersive feedback. Red line shows the acceleration measured on the platform and blue line is the acceleration calculated from the video. 
    }
    \label{fig:M3}
\end{figure}

Figure \ref{fig:M3} validates, the ability to track and replicate real-life accelerations, which compares the x and y components of acceleration across three sources: real-life skiing video, simulation environment, and the platform’s measured response. Spearman’s rank correlation coefficient $\rho$ was used to quantify similarity \cite{alihan22}. In Figure \ref{fig:M3}a, the $x$-component of acceleration from the skier video shows a strong positive correlation with the platform’s measurement $\rho = 0.8$, while the $y$-component showed $\rho = 0.65$. In Figure \ref{fig:M3}b, the simulation-based skier produces correlations of $\rho = 0.6$ for both $x$ and $y$ components. These results confirm that the platform effectively recreated real-world forces by closely matching the directional characteristics of full-body acceleration. Although exact acceleration matching was not always achievable, directional consistency was preserved.

For the skiing experiments, the extracted acceleration components directly drive the platform, as the high-dynamic motion requires explicit acceleration tracking. In contrast, for the boat experiments, roll and pitch angles are the control inputs. Because acceleration variations during boating are comparatively low, orientation-based mapping provides a sufficient approximation of inertial effects. In this case, platform tilt reorients the gravitational vector in the user’s reference frame, generating the perceived force cues without requiring direct acceleration tracking.

\subsubsection*{Multi-Material Manufacturing}\label{sec:scalability}

\begin{figure}
    \centering
        \includegraphics[width=1\textwidth]{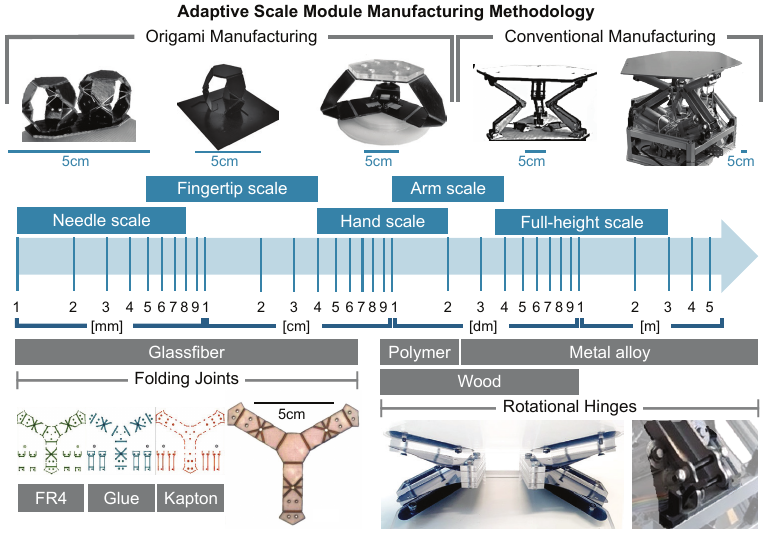}
    
    \caption{\textbf{Logarithmic scaling of the interactive robotic surface across five distinct interaction scales, from needle-scale (1–8 mm) to full-body scale (35 cm–3 m).} Material composition transitions progressively from Kapton and glass fiber composites to hybrid wood-polymer structures and industrial-grade metal frameworks. This adaptive multi-material approach ensures consistent force feedback, structural integrity, and seamless interaction fidelity across all resolutions.}
    \label{fig:M2}
\end{figure}

The platform operates across five overlapping physical scales, divided between origami-based and conventional manufacturing methods (Fig.~\ref{fig:M2}). Modules at the \textit{needle, fingertip, hand,} and small \textit{arm scale} are fabricated using origami-inspired techniques, employing folding joints made from glass fiber and Kapton sheets. Larger \textit{arm scale} and \textit{full-height scale} modules are manufacturing using conventional methods, including woodworking, metal bending, and rotational hinges. The full-height scale enables full-body immersion (FBI), providing weight support and motion-coupled force feedback.

A bottom-up, iterative design process ensures consistent performance across these scales. Small-scale prototypes are first validated for force accuracy, joint reliability, and control integration before scaling up. A hierarchical control architecture coordinates local and global actuation: small-scale modules enable fine tactile responses, while large-scale modules drive shape adaptation. This modular and scalable architecture supports diverse haptic and immersive applications.

A multi-material fabrication strategy ensures structural robustness and interaction fidelity across scales. At smaller scales, Kapton joints and glass fiber composites provide flexibility and precision. As scale increases, wood–polymer hybrids balance compliance and strength, while full-height platforms rely on aluminum structures and steel rotational joints for load-bearing tasks. To mitigate issues such as misalignment, creep, and joint failure, tolerance-aware assembly methods and material reinforcement strategies are applied throughout.

\subsection*{Ethics approval and informed consent}

All experimental protocols involving human participants were reviewed and approved by the EPFL Human Research Ethics Committee (HREC; ref. HREC000712, approved 13 March 2026). All methods involving human participants were carried out in accordance with the relevant institutional guidelines and regulations. Written informed consent was obtained from all participants before participation.

\clearpage

\textbf{Ethical approval declarations} (only required where applicable) Any article reporting experiment/s carried out on (i)~live vertebrate (or higher invertebrates), (ii)~humans or (iii)~human samples must include an unambiguous statement within the methods section that meets the following requirements: 

\begin{enumerate}[1.]
\item Approval: a statement which confirms that all experimental protocols were approved by a named institutional and/or licensing committee. Please identify the approving body in the methods section

\item Accordance: a statement explicitly saying that the methods were carried out in accordance with the relevant guidelines and regulations

\item Informed consent (for experiments involving humans or human tissue samples): include a statement confirming that informed consent was obtained from all participants and/or their legal guardian/s
\end{enumerate}

If your manuscript includes potentially identifying patient/participant information, or if it describes human transplantation research, or if it reports results of a clinical trial then additional information will be required. Please visit (\url{https://www.nature.com/nature-research/editorial-policies}) for Nature Portfolio journals, (\url{https://www.springer.com/gp/authors-editors/journal-author/journal-author-helpdesk/publishing-ethics/14214}) for Springer Nature journals, or (\url{https://www.biomedcentral.com/getpublished/editorial-policies\#ethics+and+consent}) for BMC.

\section{Conclusion}\label{sec13}

This work establishes a quantitative framework for evaluating FBI in interactive environments by defining it as geometric similarity between real and virtual motion patterns. The immersion index provides an objective alternative to subjective assessments. Because the immersion index relies on task-related submetrics rather than platform-specific parameters, it enables consistent comparison across different immersive systems and interaction modalities.

The experimental results demonstrate that essential inertial cues can drive whole-body responses even with limited mechanical degrees of freedom. This suggests that scalable immersive environments do not require full replication of real-world kinematics, but instead must preserve the key directional force characteristics that shape human motion. As a result, modular platforms such as MIROS can support physically grounded immersive interaction while maintaining architectural simplicity and scalability.

Future work will extend this framework to distributed multi-module surfaces, broader task domains with additional surface cues such as roughness to further evaluate the generalizability of the immersion index. These directions will help establish standardized benchmarks for evaluating immersive human–environment interaction across robotics, virtual environments, and teleoperation systems.

\bmhead{Supplementary information}

If your article has accompanying supplementary file/s please state so here. 

Authors reporting data from electrophoretic gels and blots should supply the full unprocessed scans for key as part of their Supplementary information. This may be requested by the editorial team/s if it is missing.

Please refer to Journal-level guidance for any specific requirements.

\bmhead{Acknowledgements}

Acknowledgements are not compulsory. Where included they should be brief. Grant or contribution numbers may be acknowledged.

Please refer to Journal-level guidance for any specific requirements.

\section*{Declarations}

\textbf{Funding} This work was supported by the Swiss National Science Foundation Fund for Approaching Tangible Reality with Reconfigurable Robots project (514770).

\textbf{Conflict of interest/Competing interests (check journal-specific guidelines
for which heading to use).} The authors declare no competing interests.

\textbf{Ethics approval and consent to participate.} All procedures performed in studies involving human users were in accordance with the ethical standards of the institutional research committee. The study (ref: HREC000712) was reviewed and approved by the EPFL Human Research Ethics Committee (HREC), on 13 March 2026.

\textbf{Consent for publication.}

\textbf{Data availability.} All data generated or analyzed during this study are included in
the published article, and are available from the corresponding author on reasonable
request.

\textbf{Materials availability.} Not applicable

\textbf{Code availability.} Not applicable

\textbf{Author contributions:} A.B., F.Z. and J.P. designed the study and interpreted the results. A.B., F.Z. and J.P. conceived the idea of full-body immersion (FBI). A.B., E.Y. and J.P. designed the experiments and A.B. and E.Y. conducted the experiments. A.B., E.Y. and F.Z. developed and build the robotic platforms. A.B., E.Y. and F.B. designed the simulation environment in UE5. A.B. designed the electronics, communication and control loop of the robots. A.B. and F.Z. produced the figures and videos. A.B., E.Y. F.Z. and J.P. wrote the manuscript.

Some journals require declarations to be submitted in a standardised format. Please check the Instructions for Authors of the journal to which you are submitting to see if you need to complete this section. If yes, your manuscript must contain the following sections under the heading `Declarations':

\begin{itemize}
\item Funding
\item Conflict of interest/Competing interests (check journal-specific guidelines for which heading to use)
\item Ethics approval and consent to participate
\item Consent for publication
\item Data availability 
\item Materials availability
\item Code availability 
\item Author contribution
\end{itemize}

\noindent
If any of the sections are not relevant to your manuscript, please include the heading and write `Not applicable' for that section. 

\bigskip
\begin{flushleft}%
Editorial Policies for:

\bigskip\noindent
Springer journals and proceedings: \url{https://www.springer.com/gp/editorial-policies}

\bigskip\noindent
Nature Portfolio journals: \url{https://www.nature.com/nature-research/editorial-policies}

\bigskip\noindent
\textit{Scientific Reports}: \url{https://www.nature.com/srep/journal-policies/editorial-policies}

\bigskip\noindent
BMC journals: \url{https://www.biomedcentral.com/getpublished/editorial-policies}
\end{flushleft}

\begin{appendices}

\section{Section title of first appendix}\label{secA1}

An appendix contains supplementary information that is not an essential part of the text itself but which may be helpful in providing a more comprehensive understanding of the research problem or it is information that is too cumbersome to be included in the body of the paper.




\end{appendices}


\bibliography{sn-bibliography}

\end{document}